\documentclass[accepted]{uai2023} % after acceptance, for a revised
\usepackage[american]{babel}

\usepackage{natbib} % has a nice set of citation styles and commands
\usepackage{mathtools} % amsmath with fixes and additions
\usepackage{booktabs} % commands to create good-looking tables
\usepackage{tikz} % nice language for creating drawings and diagrams

\usepackage{graphicx}
\usepackage{dutchcal}
\usepackage{amsmath}
\usepackage{amssymb}
\usepackage{balance}

\def\gL{{\mathcal{L}}}
\def\gG{{\mathcal{G}}}
\def\gH{{\mathcal{H}}}
\def\gN{{\mathcal{N}}}
\def\gh{{\mathcal{h}}}
\def\gi{{\mathcal{i}}}
\def\gj{{\mathcal{j}}}
\def\gm{{\mathcal{m}}}
\def\gy{{\mathcal{y}}}
\def\gg{{\mathcal{g}}}

\title{$E(2)$-Equivariant Vision Transformer}

\author[1]{\href{mailto:rux@zju.edu.cn}{Renjun~Xu}\thanks{Contributed equally.}}
\author[1,2]{\href{mailto:yangkaifan@zju.edu.cn}{Kaifan~Yang$^*$}}
\author[1,2]{\href{mailto:lk2017@zju.edu.cn}{Ke~Liu$^*$\thanks{Corresponding author: Ke Liu}}}
\author[3,2]{\href{mailto:F.He@ed.ac.uk}{Fengxiang~He}}
\affil[1]{%
    College of Computer Science and Technology\\
    Zhejiang University%\\
}
\affil[2]{%
    JD Explore Academy, JD.com, Inc.
}
\affil[3]{%
    AIAI, School of Informatics, University of Edinburgh
}

\begin{document}
\maketitle

\renewcommand{\thefootnote}{\fnsymbol{footnote}}

\begin{abstract}
Vision Transformer (ViT) has achieved remarkable performance in computer vision. However, positional encoding in ViT makes it substantially difficult to learn the intrinsic equivariance in data. To approach a general solution for all E(2) settings, we design a Group Equivariant Vision Transformer (GE-ViT) via a novel, effective positional encoding operator. We prove that GE-ViT meets all the theoretical requirements of an equivariant neural network. Comprehensive experiments are conducted on standard benchmark datasets, demonstrating that GE-ViT significantly outperforms non-equivariant self-attention networks. The code is available at \href{https://github.com/ZJUCDSYangKaifan/GEVit}{https://github.com/ZJUCDSYangKaifan/GEVit}. 
\end{abstract}

\section{Introduction}
Equivariance is an intrinsic property of many domains, such as image processing \citep{dataAug}, 3D point cloud processing \citep{li2018pointcnn}, chemistry \citep{faber2016machine}, astronomy \citep{ntampaka2016dynamical}, economics \citep{qin2022benefits}, etc. Translation equivariance is naturally guaranteed in CNNs, i.e., if a pattern in an image is translated, the learned image representation by a CNN is also translated in the same way. However, realizing equivariance is not natural for other models or groups. \citet{zaheer2017deep}, \citet{2016cohen}, and \citet{cohen2019} adopt machine learning to realize the equivariance via modifying classic neural networks. In visual tasks, the equivariance has been highlighted in the aspects of permutation~\citep{GSA-Nets}, symmetry \citep{dataAug}, and translation \citep{HarmonicNetwork}. 

Vision Transformer (ViT) \citep{vit} based on self-attention has been widely used in computer vision. According to the theoretical analyze (\S\ref{SAConclusion}), it is the positional encoding that destroys the equivariance of self-attention. To extend the equivariance of ViT to arbitrary affine groups, a new positional encoding should be designed to replace the traditional one. 
Initial attempts have been made to modify the self-attention to be equivariant \citep{GSA-Nets,se3Transformer,LieTransformer}.
The SE(3)-Transformers \citep{se3Transformer} takes the irreducible representations of SO(3) and LieTransformer \citep{LieTransformer} utilizes the Lie algebra. However, they focus on processing 3-D point cloud data. GSA-Nets \citep{GSA-Nets} proposed new positional encoding operations, which meet challenges in some cases. 

To address this issue, we propose a \textbf{G}roup \textbf{E}quivariant \textbf{Vi}sion \textbf{T}ransformer (\textbf{GE-ViT}) via a novel, effective equivariant positional encoding operation. We prove that the GE-ViT has met the theoretical requirements of a group equivariant neural network. 

The equivariance in GE-ViT brought advantages over previous works.
The group equivariance significantly improves the generalization for its equivariance on group \citep{sannai2021improved, he2020recent}.
Parameter efficiency and steerability \citep{cohen2016steerable,SFCNN} are also guaranteed. 
The weights of group equivariant CNN kernels are tied to particular positions of neighborhoods on the group, which requires a large number of parameters. While GE-ViT leverages long-range dependencies on group functions under a fixed parameter budget, which can express any group convolutional kernel \citep{GSA-Nets}. GE-ViT is steerable since group operations are performed directly on the positional encoding \citep{SFCNN}. The performance of GE-ViT is evaluated by experiments which fully support our algorithm.

The contributions of this work are summarized as follows:
\begin{itemize}
	\item We propose a novel Group Equivariant Vision Transformer (GE-ViT). Mathematical analysis demonstrates that the theoretical requirements of an equivariance neural network are met in GE-ViT.
	\item We conduct experiments on standard benchmark datasets. The empirical results demonstrate consistent improvements of GE-ViT over previous works.
\end{itemize}

The rest of this paper is organized as follows. Section \ref{S2}  reviews the related works. Section \ref{S3} introduces self-attention in detail and defines the notations in our paper. Preliminary concepts on groups and equivariance are introduced in Section \ref{S4}. Theory analysis of GE-ViT, especially that regarding positional encoding, is presented in Section \ref{S5}. We report the experiments in Section \ref{S6}. The discussion and future work are given in Section \ref{S7}.

\section{Related Work}\label{S2}
Transformer \citep{Transformer} and its variants \citep{BERT} have achieved remarkable success in natural language processing (NLP) \citep{Transformer}, computer vision (CV) \citep{carion2020end,vit,swin-vit}, and other fields \citep{alphafold2,s2snet}. Different from previous methods, e.g., recurrent neural networks (RNNs) \citep{rnn} and convolutional neural networks (CNNs) \citep{cnn}, transformer handles the input tokens simultaneously, which has shown competitive performance and superior ability in capturing long-range dependencies between these tokens. The core of transformer is the self-attention operation \citep{Transformer}, which excels at modeling the relationship of tokens in a sequence. Self-attention takes the similarity of token representations as attention scores and updates the representations with the score weighted sum of them in an iterative manner.

The group equivariant neural network was first proposed by \citet{2016cohen}, which extended the equivariance of CNNs from translation to discrete groups. The main idea of the approach is that it uses standard convolutional kernels and transforms them or the feature maps for each of the elements in the group \citep{cohen2019}. This approach is easy to implement and has been used widely \citep{marcos2017rotation,zhou2017oriented}. However, this kind of approach can only be used in particular circumstances where locations are discrete and the group cardinality is small such as image data. 

Nowadays, many methods have been proposed for designing group equivariant networks. The equivariance of networks has been extended to general symmetry groups \citep{bekkers2019b,venkataraman2019building,E22019}. Macroscopically, equivariant neural networks can be broadly categorised by whether the input spatial data is lifted onto the space of functions on group $G$ or not \citep{LieTransformer}. Without lifting, the equivariant map is defined on the homogeneous input space $\mathcal{X}$. For convolutional networks, the kernel is always expressed using a basis of equivariant functions, such as circular harmonics \citep{SFCNN,HarmonicNetwork}, spherical harmonics \citep{TFN}. With lifting, the equivariant map is defined on $G$ \citep{cohen2018spherical,esteves2018learning,LieConv,LieTransformer,romero2019co}. GE-ViT uses lifting to design equivariant self-attention \citep{GSA-Nets}.

Research on how to make the self-attention satisfy the general group equivariance is already existed \citep{romero2020attentive}. The SE(3)-Transformer \citep{se3Transformer} achieves this goal via the irreducible representations of SO(3) and LieTransformer \citep{LieTransformer} achieves this by means of Lie algebra. However, GE-ViT, the model proposed by this paper, achieved this by designing a new positional encoding. Besides, the above two models are specifically designed for processing 3-D point cloud data while GE-ViT is good at processing regular image data. 

\section{Vision Transformer}\label{S3}

In this section, we formally formulate vision transformers \citep{vit}.

\subsection{Architecture}

We first define some notations for brevity.
Set $\{1,2,3,\dots,n\}$ is denoted by $[n]$. Let $\mathcal{S}=[N]$. $L_{\mathcal{V}}(\mathcal{S})$ denote the space of functions $\left\{f:\mathcal{S}\rightarrow \mathcal{V}\right\}$, where $\mathcal{V}$ represents a vector space. Accordingly, a matrix $\textbf{X} \in \mathbb{R}^{N\times C_{\rm in}}$ can be interpreted as a vector-valued function  $f_X:\mathcal{S}\rightarrow \mathbb{R}^{C_{\rm in}}$ that maps element $i \in \mathcal{S}$ to $C_{\rm in}$-dimension vector $\textbf{X}_i\in \mathbb{R}^ {C_{\rm in}}$. A matrix multiplication, $\textbf{XW}_{y}^{\top}$ between matrices $\textbf{X} \in \mathbb{R}^{N\times C_{\rm in}}$ and $\textbf{W}_{y} \in \mathbb{R}^{C_{\rm out}\times C_{\rm in}}$ can be represented as a function $\varphi_{y}:L_{\mathbb{R}^{C_{\rm in}}}(\mathcal{S}) \rightarrow L_{\mathbb{R}^{C_{\rm out}}}(\mathcal{S})$, as $\varphi_y(f_X)=f_{XW_y^T}$.

ViT reshapes the image into a sequence of 2D tokens, which are then flattened and mapped into token embeddings, i.e. vectors, with a trainable linear projection \citep{vit}. To get the structural information of an image involved, the positional encodings are calculated and aggregated with the corresponding token embeddings to form the representations of the tokens. Finally, self-attention mechanisms are performed on these token representations.

\begin{figure}[t!]
    \centering
	\includegraphics[width=0.7\columnwidth]{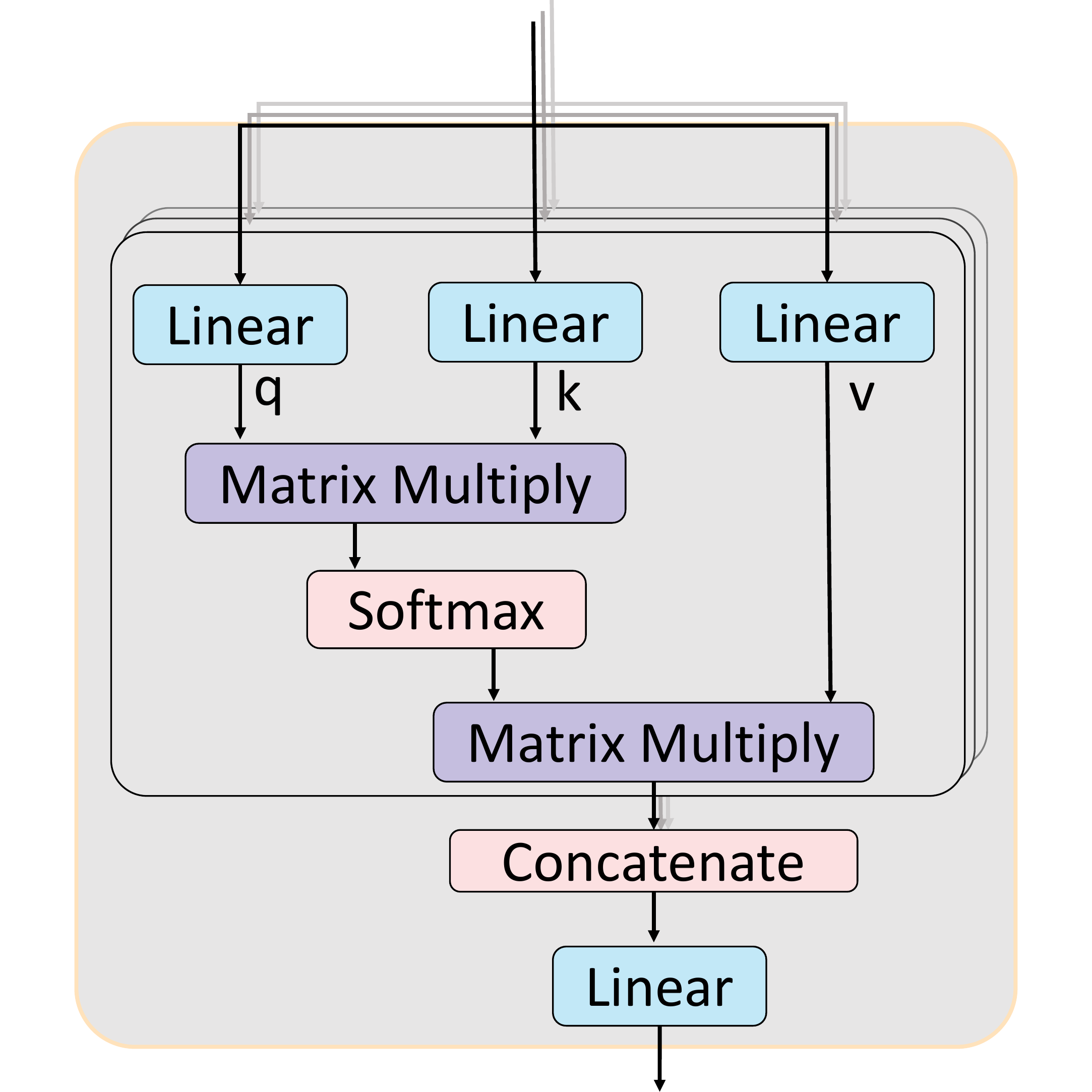}
	\caption{Illustration of self-attention. $q$, $k$, and $v$ denote the query, key, and value, respectively. ``Linear'' denotes the fully connected neural network layers. For multi-head self-attention, each black box denotes one head and gives a representation. All the representations are concatenated through the concatenate layer and input into the linear layer.
	}\label{fig:qkv}
\end{figure}

\subsection{Self-Attention} 
The overview of self-attention is shown in Fig.~\ref{fig:qkv}. A self-attention module takes in $N$ inputs and returns $N$ outputs. Let $\mathbf{X}\in \mathbb{R}^{N\times C_{\rm in}}$ be an input matrix consisting of $N$ tokens of $C_{\rm in}$ dimensions. Let $\mathbf{Y}\in \mathbb{R}^{N\times C_{\rm out}}$  be an output matrix consisting of $N$ tokens of $C_{\rm out}$ dimensions obtained from $\mathbf{X}$ through self-attention. The whole calculation process can be divided into the following two steps:
\begin{enumerate}
    \item Calculate the attention scores matrix $\mathbf{A}\in \mathbb{R}^{{N\times N}}$.
    \begin{align}\label{SWP}
        \mathbf{A}:=\mathbf{XW}_{\rm qry}(\mathbf{XW_{\rm key}})^{\top},
    \end{align}
    where $\mathbf{W}_{\rm qry},\mathbf{W}_{\rm key} \in \mathbb{R}^{C_{\rm in}\times C_{\rm h}}$ represent query and key matrices respectively. $\mathbf{A}_{i,j}$ represents the correlation between the $i$-th item and the $j$-th item of the input.
    \item Get the output through softmax and summation.
    \begin{align}\label{SA}
        \textbf{Y} = \rm{SA}(\mathbf{X}) := \rm{softmax}_{[\;,:\;]}(\mathbf{A})\mathbf{XW_{\rm val}},
    \end{align}
    where $\mathbf{W}_{\rm val} \in \mathbb{R}^{C_{\rm in}\times C_{\rm h}}$ represents value matrix.
\end{enumerate}
In practical application, Multi-Headed Self-Attention (MHSA) that focuses on different aspects of the input is applied. The outputs of different heads of dimension $C_{\rm h}$ are concatenated firstly and then projected to output via a projection matrix $\mathbf{W}_{\rm out} \in \mathbb{R}^{{HC_{\rm h}\times C_{\rm out}}}$. The $H$ denotes the number of heads.
\begin{align}\label{MHSA}
    {\rm{MHSA}}(\mathbf{X}) := \mathop{\rm{concat}}\limits_{h \in [H]}\;[{\rm{SA}}^{(h)}(\mathbf{X})]\mathbf{W}_{\rm out}.
\end{align}

According to the above, we define attention score matrix (Eq. \ref{SWP}) without positional encoding as below: 
\begin{align}
    \mathbf{A}_{i,j} = \alpha[f](i,j) = \left\langle\varphi_{\rm qry}(f(i)),\varphi_{\rm key}(f(j))\right\rangle .
\end{align}
The function $\alpha[f]: \mathcal{S} \times \mathcal{S} \rightarrow \mathbb{R}$ maps pairs of set elements $i,j \in \mathcal{S}$ to an attention score $\mathbf{A}_{i,j}$. Consequently, the self-attention (Eq. \ref{SA}) can be represented as below:
\begin{align}
    \mathbf{Y}_{i,:}&=\zeta[f](i) = \nonumber \sum\limits_{j\in\mathcal{S}}\sigma_{j}(\alpha[f](i,j))\varphi_{\rm val}(f(j)) \\
     &= \sum\limits_{j\in\mathcal{S}}\sigma_{j}(\left\langle\varphi_{\rm qry}(f(i)),\varphi_{\rm key}(f(j))\right\rangle)\varphi_{\rm val}(f(j)),
\end{align}
where $\zeta[f]:\mathcal{S} \rightarrow \mathbb{R}^{C_{\rm h}}$, $\sigma = \rm{softmax}$, and $\sigma_j = \frac{e^{zj}}{\sum_{i=1}^{N}e^{zi}}$. Similarly, the MHSA(Eq. \ref{MHSA}) can be expressed as below:
\begin{align}
    &{\rm{MHSA}}(\mathbf{X}_{i}) = m[f](i) = \nonumber \varphi_{out}(\mathop{\bigcup}\limits_{h\in [H]}\zeta^{(h)}[f](i)) \\
    &\nonumber= \label{FMHSA}\varphi_{out}(\mathop{\bigcup}\limits_{h\in [H]}\sum\limits_{j\in\mathcal{S}}\sigma_{j}(\left\langle\varphi_{\rm qry}^{(h)}(f(i)), \right. \\ 
    & \quad\quad\quad\quad\quad\quad\quad\quad\quad \left. \varphi_{\rm key}^{(h)}(f(j))\right\rangle)\varphi_{\rm val}^{(h)}(f(j))),
\end{align}
where $\cup$ is the concatenation operator and  $\mathcal{S} \rightarrow \mathbb{R}^{C_{\rm out}}$.

To handle the quadratic time complexity of the self-attention, ViT only uses the regions on an image nearest to the $i_{th}$ item, when calculating the output of the $i_{th}$ item. Let $\eta_{(i)}$ be the selected part related to the $i_{th}$ item, which is also called the local neighbourhood of the token $i$ in the later section. Therefore, replacing $\mathcal{S}$ with $\eta_{(i)}$, Eq.~\ref{FMHSA} can be written as below:
\begin{align}
    &{\rm{MHSA}}(\mathbf{X}_{i}) = m[f](i) = \nonumber \varphi_{out}(\mathop{\bigcup}\limits_{h\in [H]}\zeta^{(h)}[f](i)) \\
    &\nonumber= \varphi_{out}(\mathop{\bigcup}\limits_{h\in [H]}\sum\limits_{j\in\eta{(i)}}\sigma_{j}(\left\langle\varphi_{\rm qry}^{(h)}(f(i)),\right.\\
    & \quad\quad\quad\quad\quad\quad\quad\quad\quad \left.\varphi_{\rm key}^{(h)}(f(j))\right\rangle)\varphi_{\rm val}^{(h)}(f(j))).
\end{align}

\subsection{Positional Encoding}

The self-attention overlooks structural information.
To solve this issue, positional encoding $\mathbf{P}$ is proposed \citep{vit}, as introduced below.

\paragraph{Absolute Positional Encoding}
In absolute positional encoding, every position is given a unique positional encoding. The whole positional encoding can be represented by a matrix $\mathbf{P}\in \mathbb{R}^{N\times C_{\rm in}}$. Consequently, the attention scores matrix $\mathbf{A}$ can be formulated as follows:
\begin{align}\label{APE}
    \mathbf{A}:=\mathbf{(X+P)W}_{\rm qry}(\mathbf{(X+P)W_{\rm key}})^{\top}.
\end{align}
The positional encoding is a function $\rho: \mathcal{S} \rightarrow \mathbb{R}^{C_{\rm in}}$ that maps set elements $i \in \mathcal{S}$ to a vector representation. Using this definition, Eq. \ref{APE} can be written as:
\begin{align}
    \nonumber m[f,p](i) = \varphi_{out}&(\mathop{\cup}\limits_{h\in [H]}\sum\limits_{j\in\eta(i)}\sigma_{j}(\left\langle\varphi_{\rm qry}^{(h)}(f(i)+\rho(i)), \right. \\
    &\left. \varphi_{\rm key}^{(h)}(f(j)+\rho(j))\right\rangle)\varphi_{\rm val}^{(h)}(f(j)))
\end{align}

\paragraph{Relative Positional Encoding} Proposed by \citet{RE}, relative positional encoding considers the relative distance between the query token $i$ and the key token $j$. The corresponding attention score $\mathbf{A}_{i,j}$ can be calculated by the following formula:
\begin{align}\label{RPE}
    \mathbf{A}_{i,j}^{\rm rel}:=\mathbf{X}_i\mathbf{W}_{\rm qry}((\mathbf{X}_{j}+\mathbf{P}_{x(j)-x(i)})\mathbf{W}_{\rm key})^{\top},
\end{align}
where $x(i)$ is the position of token $i$, and $\mathbf{P}_{x(j)-x(i)} \in \mathbb{R}^{1 \times C_{\rm in}}$ is the positional encoding of the relative distance of token $i$ and token $j$.
Similarly, relative positional encoding can be defined as $\rho(i,j)=\rho^{P}(x(j)-x(i)) $
of pairs $(i,j), i\in \mathcal{S}, j\in \eta(i)$.
Thus, the Eq. \ref{RPE} can be written as:
\begin{align}
    \nonumber m[f,p](i) = \varphi_{out}(\mathop{\bigcup}\limits_{h\in [H]}\sum\limits_{j\in\eta(i)}\sigma_{j}(\left\langle\varphi_{\rm qry}^{(h)}(f(i)), \right. \\
    \varphi_{\rm key}^{(h)}(f(j) +\rho(i,j))\Big\rangle)\varphi_{\rm val}^{(h)}(f(j)))
\end{align}

\section{Group Equivariance}\label{S4}

This section presents necessary definitions and notations in group representation theory and group equivariance properties.

\subsection{Group Representation Theory}

\paragraph{Group} 
The group is an abstract mathematical concept. Formally a group $(G; \circ)$ consists of a set $G$ and a binary composition operator $\circ: G \times G \rightarrow G$. All groups must adhere to the following 4 axioms:
\begin{enumerate}
    \item Closure: $g\circ h \in G$ for all $g,h,\in G$.
    \item Associativity: $f \circ (g \circ h)$ = $(f \circ g) \circ h$ = $f \circ g \circ h$  for all $f, g, h \in G$.
    \item Identity: There exists an element  such that $e \circ g$ = $g \circ e = g $ for all $g \in G$.
    \item Inverses: For each $g \in G$ there exists a $g^{-1} \in G$ such that $g^{-1} \circ g = g \circ g^{-1}= e$.
\end{enumerate}
Each group element $g \in G$ corresponds to a symmetry transformation. In practice, the binary composition operator $\circ$ can be omitted. 
Groups can be finite or infinite, countable or uncountable, compact or non-compact. 
Note that they are not necessarily commutative; that is, $gh \neq hg$ in general. If a group is commutative, that is $gh = hg$ for all $g,h \in G$, it is called the Abelian Group. One example of the infinite group is $E(2)$, the set of all 2D rotations about the origin and the 2D translation. Because the image is transformed in 2D, $E(2)$ is the focus of this paper.

\paragraph{Group Action}
A group action $\rho(g)$ is a bijective map from a space into itself: $\rho(g): \mathcal{X} \rightarrow \mathcal{X}$. It is parameterized by an element $g$ of a group $G$. For $\rho(g)x$, we say that $\rho(g)$ acts on $x$. A symmetry transformation of group element $g \in G$ on object $x \in \mathcal{X}$ is referred to as the group action of $G$ on $\mathcal{X}$. $\rho(g)x$ is often written as $gx$ to reduce clutter. In the context of group equivariant neural networks, grouping action object $\mathcal{X}$ is commonly defined to be the space of scalar-valued functions or vector-valued functions on some set $\mathcal{E}$, so that $\mathcal{X} = \{f | f : \mathcal{E} \rightarrow \mathbb{R}^{d}\}$. This set $\mathcal{E}$ could be a Euclidean space, e.g., a grey-scale image can be expressed as a feature map $f : \mathbb{R}^2 \rightarrow \mathbb{R}$ from pixel coordinate $x_i$ to pixel intensity $f_i$, supported on the grid of pixel coordinates.

\paragraph{Group Representation}
A group representation $\rho : G \rightarrow GL(N)$ is a map from a group $G$ to the set of $N \times N$ invertible matrices $GL(N)$.
Critically, $\rho$ is a group homomorphism, i.e., it satisfies the following property:
\begin{align}
    \rho(g1 \circ g2) = \rho(g1)\rho(g2), \quad \forall g1, g2 \in G.
\end{align}
For $SO(2)$, the standard rotation matrix is an example of a representation that acts on $\mathbb{R}^{2}$:
$$
\rho(\theta) = 
\begin{bmatrix}
    {\rm{cos}}\theta & {\rm{-sin}}\theta \\
    {\rm{sin}}\theta & {\rm{cos}}\theta
\end{bmatrix}.
$$
Accordingly, the rotation of the image can be expressed as a representation of $SO(2)$ by extending the action $\rho$ on the pixel coordinates $x$ to a representation $\pi$ that acts on the space of feature maps $\{f | f : \mathcal{E} \rightarrow \mathbb{R}^{d}\}$: 
$$
[\pi(g)(f)](x) \triangleq  f(\rho (g^{-1})x),
$$ 
where $\mathcal{E} = \{x_{i}\}$. We can write $gx$ instead of $\rho(g)x$ to reduce clutter: 
$$
[\pi(g)(f)](x) \triangleq  f(g^{-1}x).
$$
And it is equivalent to the mapping: 
$$
(x_i,f_i)_{i=1}^{n} \rightarrow (\rho(g)x_{i}, f_{i})_{i=1}^{n},
$$ 
where $n$ is the total number of pixels in the image.

\paragraph{Affine Group}\label{affineGroup}
Affine groups have the following form: $\mathcal{G} = \mathbb{R}^{d} \rtimes \mathcal{H}$. It is resulted from the semi-direct product ($\rtimes$) between the translation group ($\mathbb{R}^{d}, +$) and an group $\mathcal{H}$ that acts on $\mathbb{R}^{d}$. $\mathcal{H}$ can be rotation, mirroring, etc.

\paragraph{Group Equivariance}
 A map $\Phi: V_{1} \rightarrow V_{2}$ is $G$-equivariant with respect to actions $\rho1$, $\rho2$ of $G$ acting on $V_{1}, V_{2}$ respectively if: 
\begin{align}
    \Phi[\rho1(g)f] = \rho2(g)[\Phi[f]], \quad \forall g \in G, f \in V_{1}.
\end{align}
As is well-known, convolution is an equivariant map for the translation group.
\begin{figure}[t!]
    \centering
	\includegraphics[width=0.9\columnwidth]{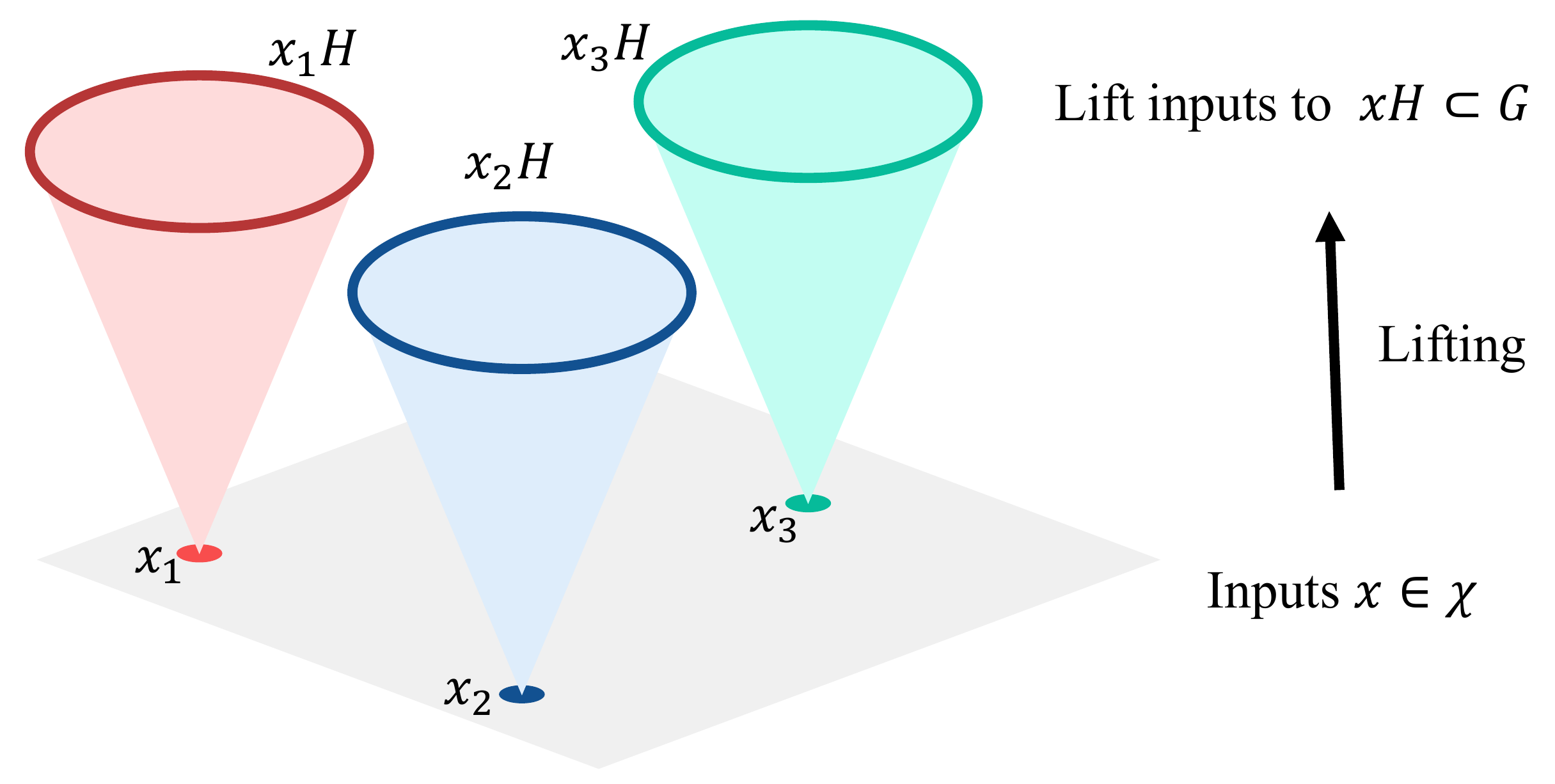}
	\caption{The illustration of lifting. For any $x\in \mathcal{X}$, $f(x)$ equals to $\mathcal{L}(f)(g)$ on $G$, where $g\in xH$, $\mathcal{L}$ is the lifting operation, and $f$ is a function defined on $\mathcal{X}$.
	}\label{fig:lift}
\end{figure}

\paragraph{Lifting}
We can view $\mathcal{X}$ as a quotient group $G/H$ for some subgroup $H$ of a group $G$, which means $\mathcal{X}$ is isomorphic to $G/H$. Then, naturally, the function $f$ defined on $\mathcal{X}$ can be viewed as defined on $G/H$. Thus, we define the lifting operation $\mathcal{L}$ on the function $f$ as below,
\begin{equation*}
    \mathcal{L}(f)(g)=f([g]),
\end{equation*}
where $[g]\in G/H$ is the equivalent class of $g$. 
For example, $\mathbb{R}^{2}$ is isomorphic to $SE(2)/SO(2)$, and every element $g\in SE(2)$ can be written as $tr$ uniquely, where $t\in \mathbb{R}^2$ and $r\in SO(2)$. Furthermore, for any function $f$ on $\mathbb{R}^2$, the lifting function $\mathcal{L}(f)$ is defined as $\mathcal{L}(f)(g)=f(t)$.

\paragraph{Equivariance of Self-Attention}\label{SAConclusion}
The proved important results on the equivariance of self-attention are as follows \citet{GSA-Nets}:
\begin{enumerate}
    \item The global self-attention formulation without positional encoding (Eq.~\ref{MHSA}) is permutation equivariant.
    \item Absolute position-aware self-attention (Eq.~\ref{APE}) is neither permutation nor translation equivariant.
    \item Relative position-aware self-attention (Eq.~\ref{RPE}) is translation equivariant.
\end{enumerate}

Our model covers $SE(2)$- and $E(2)$-equivariance, which correspond to (1) translational and rotational equivariance, and (2) translational, rotational, and reflection equivariance, respectively.

\section{Group Equivariant Vision Transformer}\label{S5}

To design an equivariant network, there are usually two choices of group representation: irreducible representation and regular representation. The experimental results~\citep{se3Transformer, LieTransformer, E22019} show that regular representation is more expressive and \citet{ReTheory} has theoretically proved it. A lifting self-attention layer is an essential module to obtain feature representation based on regular representation. The main function of the lifting layer is mapping $f_{\mathcal{X}}$ (a function defined on $\mathbb{R}^{d}$) to $\mathcal{L}[f_{\mathcal{X}}]$ (a function defined on $G$). After the lifing layer, the feature has been defined on the group $G$, which brings practical implementation problems that the group $G$ is infinite. The summation over group elements $g \in G$ is an essential step. Fortunately, extensive experiments \citep{E22019} have shown that networks using regular representations can achieve satisfactory results via proper discrete approximations.

\subsection{Lifting Self-Attention}

As previously mentioned, the lifting self-attention is a map from functions on $\mathbb{R}^{d}$ to functions on $\mathcal{G}$ and can be expressed as:
$\mathcal{m}^{r}_{\mathcal{G}\uparrow}[f, \rho]: L_{\mathcal{V}}(\mathbb{R}^{d}) \rightarrow L_{\mathcal{V}'}(\mathcal{G})$, where $\mathcal{G}$ is an affine group and $\mathcal{G} = \mathbb{R}^{d} \rtimes \mathcal{H}$. The action of group element $\mathcal{h} \in \mathcal{H}$ on relative positional encoding $\rho(i,j)$ is defined as:
$\{\mathcal{L}_{\mathcal{h}}[\rho](i, j)\}_{\mathcal{h}\in\mathcal{H}}$, $\mathcal{L}_{\mathcal{h}}[\rho](i, j) = \rho^{P}(\mathcal{h}^{-1}x(j) - \mathcal{h}^{-1}x(i)).$ Consequently, the formula of lifting self-attention can be expressed as:
\begin{align}\label{LSA}
&\mathcal{m}^{r}_{\mathcal{G}\uparrow}[f,\rho](i,\mathcal{h})=  \;\nonumber \mathcal{m}^{r}[f,\mathcal{L}_{\mathcal{h}}[\rho]](i) \\ \nonumber 
    &=\;\varphi_{out}(\mathop{\bigcup}\limits_{h\in[H]}\sum\limits_{j\in\mathcal{\eta}(i)}\sigma_{j}(\left \langle \varphi_{\rm qry}^{(h)}(f(i)),\varphi_{\rm key}^{(h)}(f(j) +\right.\\ 
    & \quad\quad\quad \mathcal{L}_{\mathcal{h}}[\rho](i,j)) \Big \rangle)\varphi_{\rm val}^{(h)}(f(j))).
\end{align}
It has been proven that the lifting self-attention defined above is equivariant to the affine group $\mathcal{G}$ \citep{GSA-Nets}.

\subsection{Group Self-Attention}\label{section:GSA}

After the lifting self-attention layer, the feature map can be viewed as a function defined on $\mathcal{G}$. So the action of group elements $\mathcal{h}\in\mathcal{H}$ on relative positional encoding $\rho(i,j)$ is defined as: $\{\mathcal{L}_{\mathcal{h}}[\rho]((i,\tilde{\mathcal{h}}), (j,\hat{\mathcal{h}}))\}_{\mathcal{h}\in\mathcal{H}}$. 
Similar to the lifting self-attention layer, the formula of group self-attention can be expressed as: 
\begin{align}\label{GSA}
    &\mathcal{m}^{r}_\mathcal{G}[f,\rho](i,\mathcal{h}) = 
    \sum_{\tilde{\mathcal{h}}\in\mathcal{H}}\mathcal{m}^{r}[f,\mathcal{L}_{\mathcal{h}}[\rho]](i,\tilde{h})
    \\ &= \;\varphi_{out}(\mathop{\bigcup}\limits_{h\in[H]}\sum\limits_{\tilde{\mathcal{h}}\in\mathcal{H}}\sum\limits_{(j,\hat{\mathcal{h}})\in\mathcal{\eta}(i,\tilde{\mathcal{h}})}\sigma_{j,\hat{\mathcal{h}}}(\left \langle \varphi_{\rm qry}^{(h)}(f(i,\tilde{\mathcal{h}})),\right.\nonumber\\ 
    & \quad\quad\left. \varphi_{\rm key}^{(h)}(f(j,\hat{\mathcal{h}}) +\mathcal{L}_{\mathcal{h}}[\rho]((i,\tilde{\mathcal{h}}),(j,\hat{\mathcal{h}}))) \right \rangle)\varphi_{\rm val}^{(h)}(f(j,\hat{\mathcal{h}}))). 
\end{align}
%However, we prove the group self-attention using the positional encoding defined as Eq. \ref{mistakePE} is not $\mathcal{G}$-equivariant. That is, 
%$$\mathcal{m}^{r}_\mathcal{G}[\mathcal{L}_{\mathcal{g}}[f],\rho](i,\mathcal{h})\neq \mathcal{L}_{\mathcal{g}}[\mathcal{m}^{r}_\mathcal{G}[f,\rho]](i,\mathcal{h}),\quad\mathcal{g}\in \mathcal{G}.$$
%Appendix~\ref{Appendix:mistake} shows the detailed proof process. 
In order to make the module satisfy the equivariant property, we propose a novel positional encoding:
\begin{align}\label{correctPE}
    \rho((i,\tilde{\mathcal{h}}), (j,\hat{\mathcal{h}})) = \rho^{P}(x(j)-x(i), \tilde{\mathcal{h}}\hat{\mathcal{h}}^{-1}\tilde{\mathcal{h}}).
\end{align}
Correspondingly, the group action on relative positional encoding can be expressed as:
$$
    \mathcal{L}_{\mathcal{h}}[\rho]((i,\tilde{h}),(j,\hat{h})) = \rho^{P}(\mathcal{h}^{-1}(x(j)-x(i)),\mathcal{h}^{-1}(\tilde{\mathcal{h}}\hat{\mathcal{h}}^{-1}\tilde{\mathcal{h}})).
$$
It can be proven (Appendix A) that using the modified version of positional encoding (Eq. \ref{correctPE}), the group self-attention is $\mathcal{G}$-equivariant. That is,
$$\mathcal{m}^{r}_\mathcal{G}[\mathcal{L}_{\mathcal{g}}[f],\rho](i,\mathcal{h})= \mathcal{L}_{\mathcal{g}}[\mathcal{m}^{r}_\mathcal{G}[f,\rho]](i,\mathcal{h}),\quad\mathcal{g}\in \mathcal{G}.$$

\subsection{GE-ViT}
\begin{figure}[t!]
    \centering
	\includegraphics[width=0.95\columnwidth]{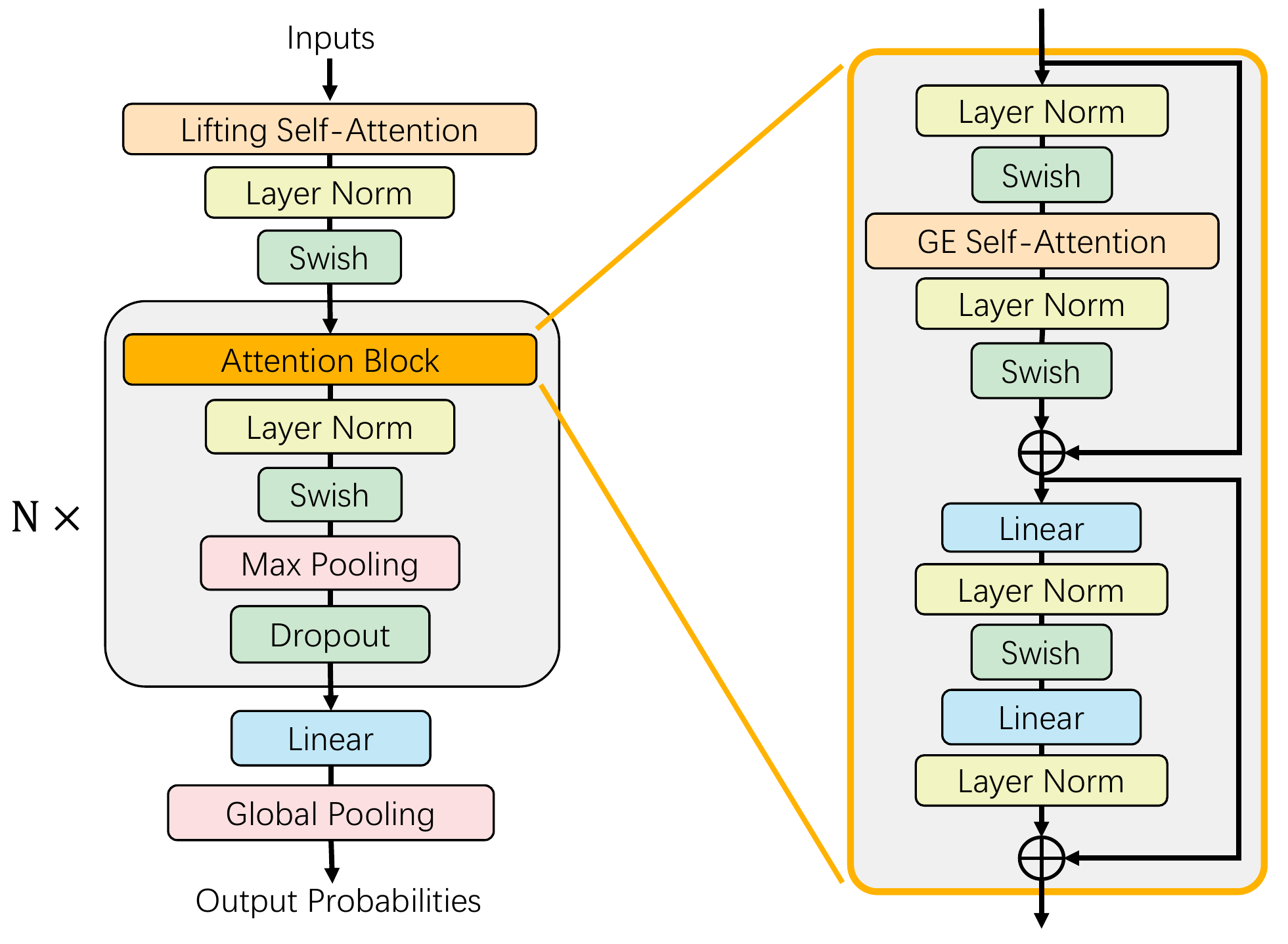}
	\caption{Illustration of GE-ViT and Attention Block. The blocks on the left show the structure of GE-ViT. Functions are transformed from R2 to Group through Lifting Self-Attention. N denotes the number of blocks in the black box. The Global Pooling block consists of max-pool over group elements followed by spatial mean-pool. Swish is an activation function \citep{swish}. The flow on the right illustrates the structure of the Attention Block. Linear denotes the fully connected neural network layers. GE Self-Attention contains lifting self-attention and group self-attention.
	}\label{fig:model}
\end{figure}

Fig.~\ref{fig:model} shows the structure of our GE-ViT. The core modules of the GE-ViT are the lifting self-attention and group self-attention \citep{GSA-Nets}. Linear map, layer normalization, and activation function are interspersed in the model. The Global Pooling block, in the end, consists of max-pool over group elements followed by spatial mean-pool \citep{vit,swin-vit,GSA-Nets}. In our experiments, we choose the local self-attention because of the computational constraints. The neighborhood size $n \times n$ denotes the chosen size of the local region. Rotation equivariant models are notated as \textbf{Rn}, where \textbf{n} represents the angle discretization. Specifically speaking, $\rm{\mathbf{R4\_SA}}$ depicts a model equivariant to rotations by $90$ degrees and $\rm{\mathbf{R8\_SA}}$ depicts a model equivariant to rotations by $45$ degrees.

\section{Experiments} \label{S6}
We conduct a study on standard benchmark datasets, rotMNIST, CIFAR-10, and PATCHCAMELYON to evaluate the performance of GE-ViT. 

\subsection{Experiment Setup}

\paragraph{Dataset} RotMNIST dataset is constructed by rotating the MNIST dataset. It is a classification dataset often used as a standard benchmark for rotation equivariance \citep{E22019}. RotMNIST contains 62,000 gray-scale $28 \times 28$ uniformly rotated handwritten digits. The rotMNIST has been divided into training, validation, and test sets of 10k, 2k, and 50k images. CIFAR-10 dataset \citep{krizhevsky2009learning} consists of 60,000 real-world $32 \times 32$ RGB images uniformly drawn from 10 classes. PATCHCAMELYON dataset \citep{veeling2018rotation} includes 327,000 $96 \times 96$ RGB image patches of tumorous/non-tumorous breast tissues.

\paragraph{Compared Approaches} 
We mainly compare our GE-ViT with $\rm{\mathbf{Z2\_SA}}$ and \textbf{GSA-Nets} \citet{vit,GSA-Nets}. $\rm{\mathbf{Z2\_SA}}$ is a translation equivariant self-attention model. \textbf{GSA-Nets} is also a self-attention-based model, which tried to introduce more kinds of equivariance to $\rm{\mathbf{Z2\_SA}}$. The reported performance of GSA-Nets is reproduced from the official released code (\href{https://github.com/dwromero/g\_selfatt}{GSA-Nets}).

\subsection{Implementation Details}

This section gives the implementation details of the experiments.

\paragraph{Invariant Network} The invariant network is a special case of the equivariant network. 
The function composited of several equivariant functions followed by an invariant function $f$, is an invariant function \citep{LieTransformer}. Therefore, the Gobal Pooling layer, an invariant map, is added to the end of the GE-ViT in our experiments.

\paragraph{Hyperparameters Setting} 
To ensure fairness, the hyperparameters remain fixed for all experiments. The number of epochs is 300 and the batch size is 8. The learning rate is set to 0.001 and the weight decay is set to 0.0001. Attention dropout rate and value dropout rate are both set to 0.1. Adam optimizer is applied.

\begin{table}[t!]
	\caption{Classification accuracy (\%) of $\rm{R4\_SA}$ with different neighborhood size on rotMNIST.}
	\centering
        \begin{tabular}{ccc}
		\toprule
		MODEL & GSA-Nets & GE-ViT (ours)\\
  \hline
  $3\times 3$ & 96.28 & \textbf{96.63} \\

  $5\times 5$ & 97.47 & \textbf{97.58} \\

 $ 7\times 7$ & 97.33 & \textbf{97.45} \\

  $9\times 9$ & 97.10 & \textbf{97.15} \\

  $11\times 11$ & 97.06 & \textbf{97.16} \\

  $15\times 15$ & 96.89 & \textbf{97.12} \\

  $19\times 19$ & 96.86 & \textbf{97.37} \\

  $23\times 23$ & 96.90 & \textbf{97.01} \\
  
		\bottomrule
	\end{tabular}
	\label{tab:1}
\end{table}

\begin{table}[t!]
    \centering
	\caption{Classification accuracy (\%) of different equivariant models on rotMNIST. All architectures based on self-attention use $5\times 5$ neighborhood size.}
        \begin{tabular}{ccc}
		\toprule
		MODEL & GSA-Nets & GE-ViT (ours)\\
  \hline
  $\rm{Z2\_SA}$ & \multicolumn{2}{c}{96.63} \\
  \hline
  $\rm{R4\_SA}$ & 97.46 & \textbf{97.58} \\

  $\rm{R8\_SA}$ & 97.79 & \textbf{97.88} \\

  $\rm{R12\_SA}$ & 97.97 & \textbf{98.01} \\

  $\rm{R16\_SA}$ & 97.66 & \textbf{97.83} \\
  
		\bottomrule
	\end{tabular}
	\label{tab:2}
\end{table}

\begin{table}[t!]
	\caption{Classification accuracy (\%) on the PATCHCAMELYON dataset.}
	\centering
        \begin{tabular}{ccc}
		\toprule
		MODEL & GSA-Nets & GE-ViT (ours)\\
  \hline
    Z2\_SA (ViT) & 80.14 & 80.14 \\
    R4\_SA  &  79.40  &  82.73 \\
    R8\_SA  & 82.26 & 83.82 \\
  
		\bottomrule
	\end{tabular}
	\label{tab:PATCHCAMELYON}
\end{table}

\subsection{Experiments and Results}
Experiments are conducted to compare our GE-ViT with previous methods. Table~\ref{tab:1} shows the classification results of $\rm{\mathbf{R4\_SA}}$ with different neighborhood size. Table~\ref{tab:2} show the classification results of different equivariant models with $5\times 5$
neighborhood size. Table~\ref{tab:PATCHCAMELYON} shows the classification results on PATCHCAMELYON dataset. The classification results of GE-ViT and GSA-Nets on CIFAR-10 are 70.40\% and 69.31\% respectively.

It is clearly observed from the experimental results that our GE-ViT outperforms other methods consistently in all settings. With more kinds of equivariance, GSA-Nets beats Z2\_SA on most settings. Our novel positional encoding improves the classification accuracy in GE-ViT and makes a new state-of-the-art. Besides, $\rm{\mathbf{R4\_SA}}$ with the neighborhood size of $5\times 5$ achieves the best accuracy. This finding is consistant with previous works \citep{GSA-Nets}. Since in the whole experiment, only the positional encodings are different and the rest remains the same, the experimental results can demonstrate the superiority of the positional encoding we proposed. The results clearly show that our GE-ViT significantly outperforms existing methods.

\begin{figure}[t!]
    \centering
	\includegraphics[width=0.9\columnwidth]{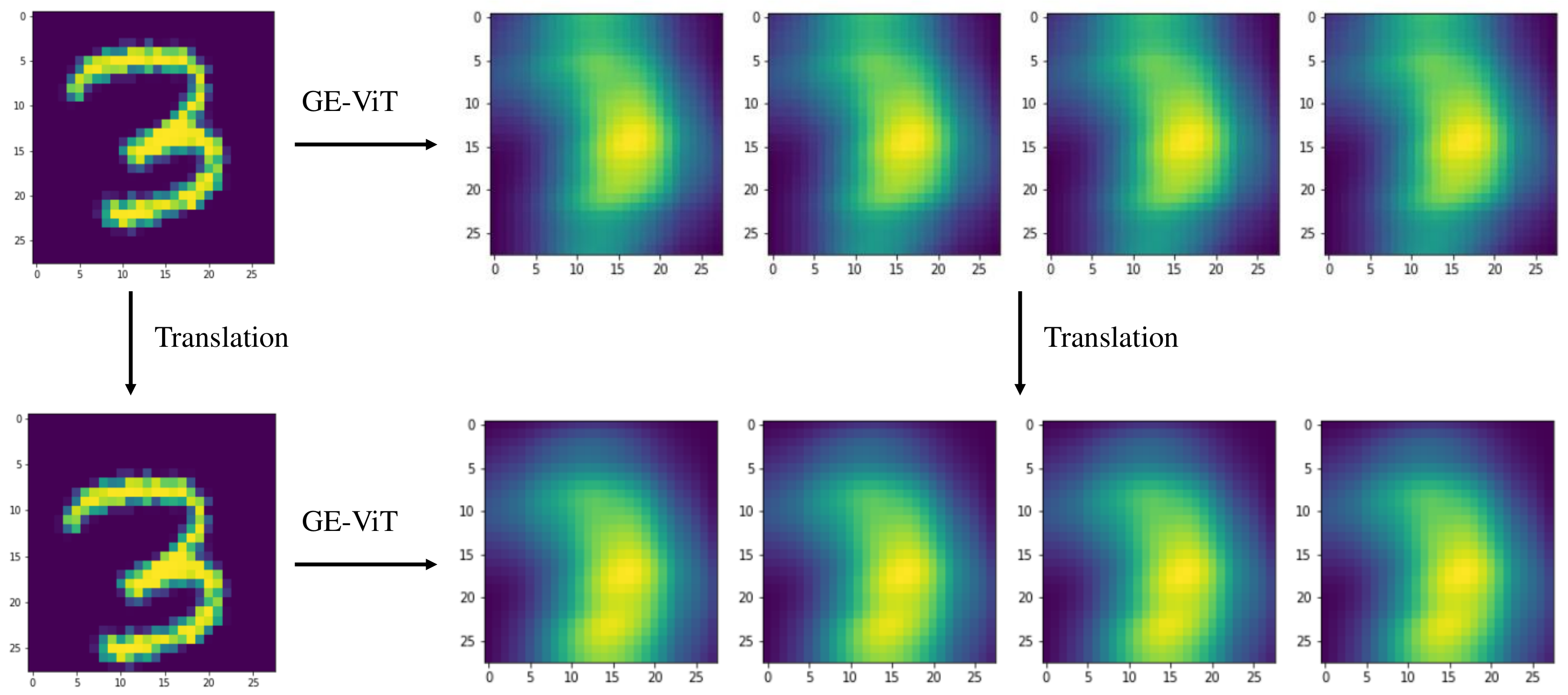}
	\caption{ Translation equivariance of GE-ViT. The images on the left are the raw data and the images on the right are feature representations. Specifically speaking, feature representations of the original data are shown in the top right of the image, and feature representations obtained by translating the original data are in the lower right of the image.  
	}\label{fig:translation}
\end{figure}

\begin{figure}[t!]
    \centering
	\includegraphics[width=0.9\columnwidth]{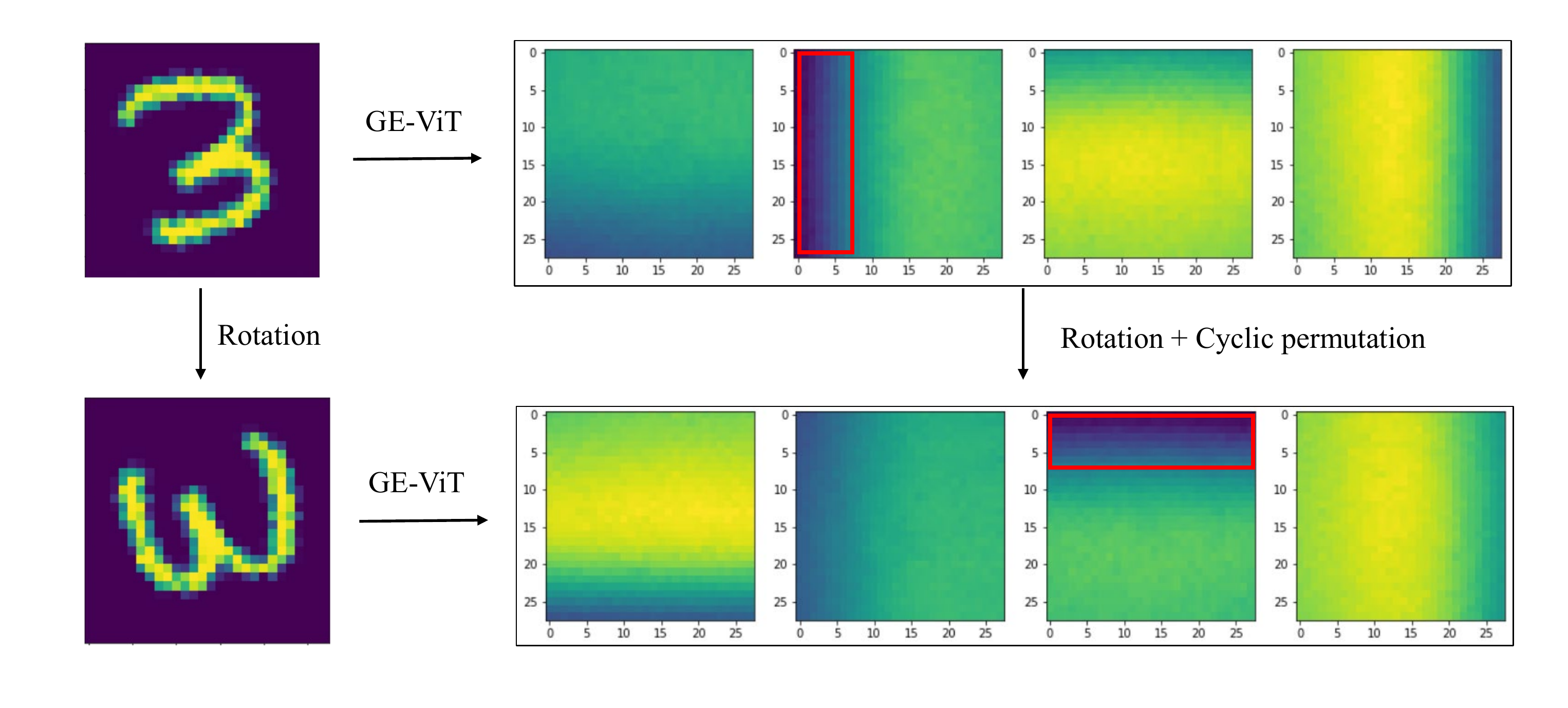}
	\caption{Rotation equivariance of GE-ViT. The images on the left are the raw data and the images on the right are feature representations. Specifically speaking, feature representations of the original data are shown in the top right of the image, and feature representations obtained by rotating the original data are in the lower right of the image.
	}\label{fig:rotation}
\end{figure}

\begin{figure}[t!]
    \centering
	\includegraphics[width=1\columnwidth]{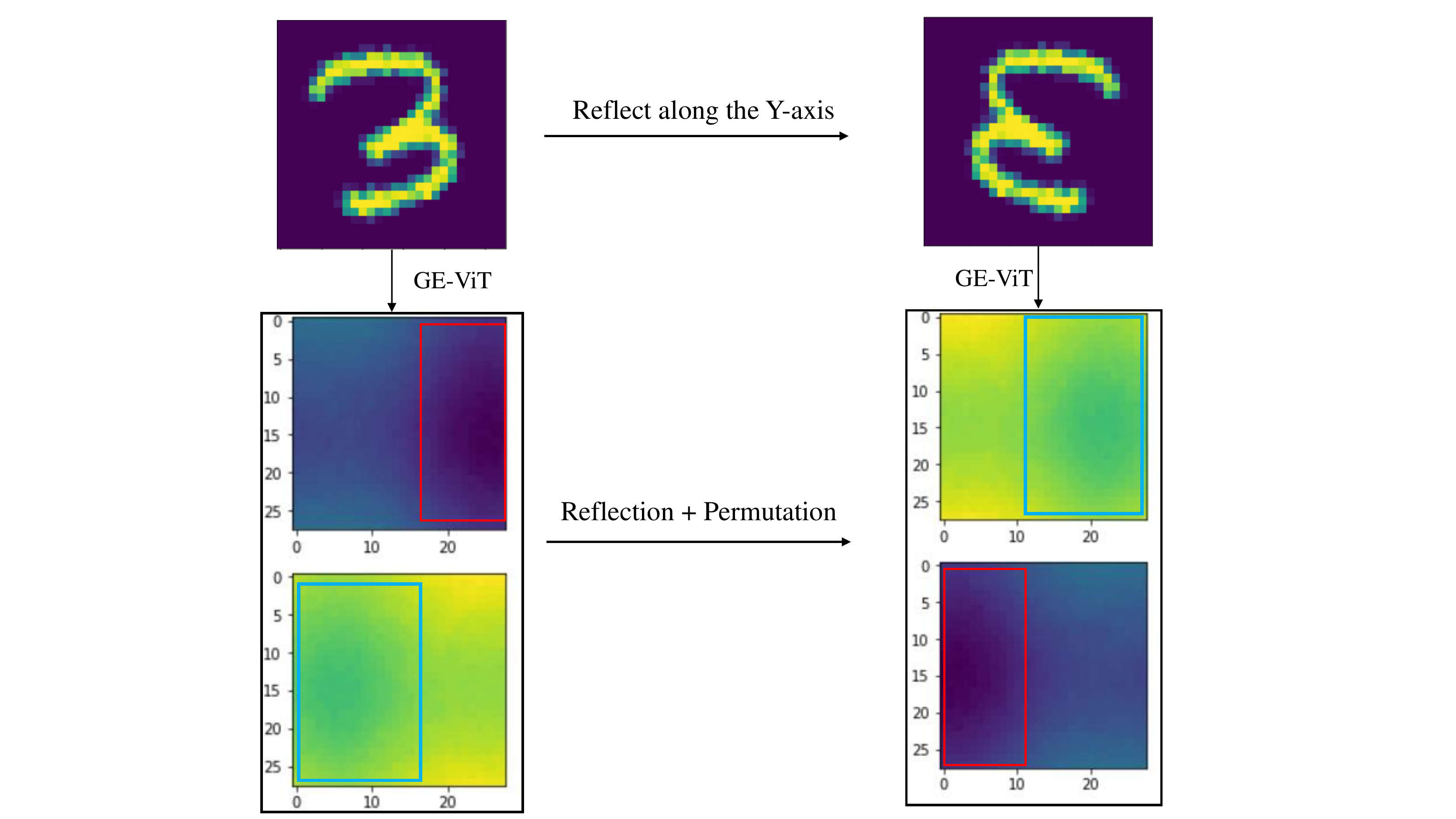}
	\caption{Reflection equivariance of GE-ViT. The images on the top are the raw data and the images on the bottom are feature representations. Specifically speaking, feature representations of the original data are shown in the lower left of the image, and feature representations obtained by flipping the original data are in the lower right of the image.  
	}\label{fig:reflection}
\end{figure}

The translation, rotation, and reflection equivariances of our GE-ViT are shown visually in Fig.~\ref{fig:translation}, Fig.~\ref{fig:rotation}, and Fig.~\ref{fig:reflection} respectively.

\begin{table}[t!]
	\caption{Comparison with equivariant convolutional networks on rotMNIST}
	\centering
	\begin{tabular}{rcrc}
		\toprule
 Model & ACC(\%) & Model & ACC(\%) \\
Z2\_SA & 96.63\% & R16\_SA & 97.83\% \\
R4\_SA & 97.58\% & Z2-CNN & 95.14\% \\
R8\_SA & 97.88\% & P4-CNN & 98.21\% \\
R12\_SA & 98.01\% & $\alpha$-P4-CNN & 98.31\% \\
 		\bottomrule
	\end{tabular}
	\label{tab:cnn}
\end{table}

GE-ViT is also compared with equivariant convolutional networks. We compare our GE-ViT with classic convolutional networks (Z2CNN) \cite{2016cohen} and equivariant convolutional networks that incorporate attention mechanisms (P4-CNN, Alpha-P4-CNN) \citep{romero2020attentive}. The size of the convolutional kernel is 3, and the settings for the other hyperparameters follow the original paper. The experimental results on the rotMNIST dataset are shown in the Table~\ref{tab:cnn}, from which, we can draw two conclusions:
\begin{itemize}[leftmargin=*]
    \item Z2\_SA performs better than Z2CNN, which demonstrates the potential of equivariant attention networks on image classification tasks. This is consistent with the conclusion in previous works that attention-based equivariant networks theoretically outperform convolution-based equivariant networks \citep{GSA-Nets}.
    \item Although our GE-ViT achieves comparable performance with equivariant convolutional networks, there is still a slight gap between them. The reasons are as follows: Firstly, the number of parameters for GE-ViT is approximately 45,000, while the number of parameters for G-CNN is around 75,000. The smaller number of parameters limit the expressiveness of the model. %Secondly, the optimization in GE-ViT are significantly different from CNN, thus the carefully crafted initialization and optimization procedures developed for CNNs over the years. Specifically, compared to convolutional networks, 
    Secondly, although GE-ViT is theoretically superior, it is more difficult to optimize \citep{liu2020understanding,zhao2020exploring}. With further research on optimization issues related to attention mechanisms, the performance of GE-ViT would gain a significant improvement.
\end{itemize}
%Following the suggestions of reviewers, w
We also conducted additional experiments to compare our GE-ViT with CPVT \citep{chu2021conditional}, which proposed a novel positional encoding method. The accuracy of CPVT on RotMNIST is 97.69\% which is worse than our GE-ViT since the positional encoding in CPVT is not equivariant.

\begin{figure}[t!]
    \centering
	\includegraphics[width=0.9\columnwidth]{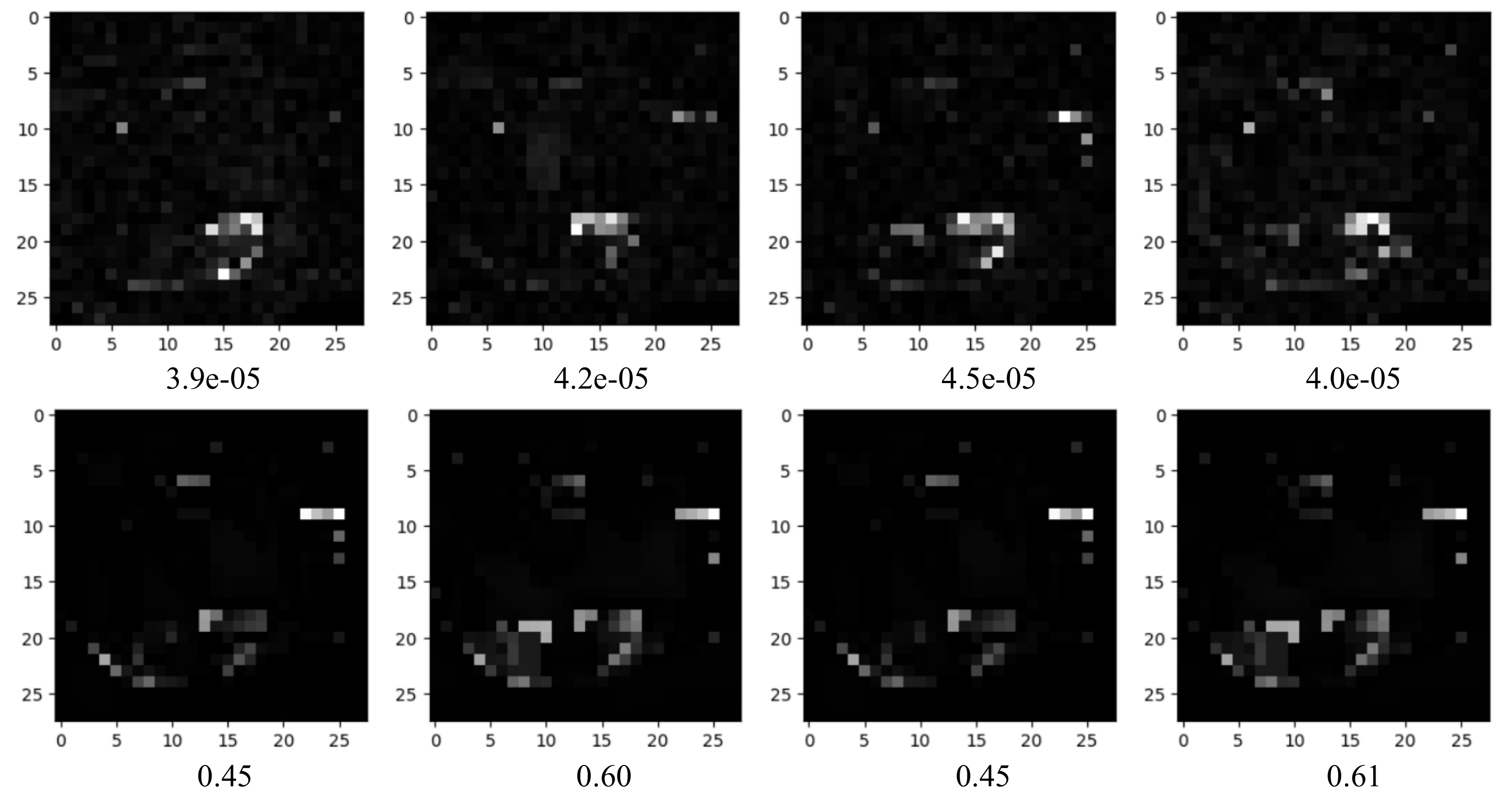}
	\caption{Error maps of GE-ViT and GSA-Net. The numbers between under the image are the average error. Images in the first and second rows are the error maps of GE-ViT and GSA-Net respectively.
	}\label{fig:errormap}
\end{figure}

Fig.~\ref{fig:errormap} shows the visual comparison between GSA-Net and GE-ViT. We visualize the errors of the feature maps in GE-ViT and GSA-Nets. Like Fig.~\ref{fig:rotation}, we visualize the error between corresponding feature maps. The process of our visualization is as follows: 
(1)	Given an image, we extracted feature maps $F_{GE}$ and $F_{GSA}$ from GE-ViT and GSA-Nets respectively.
(2)	Then, we rotated the image and extracted feature maps $F'_{GE}$ and $F'_{GSA}$ from GE-ViT and GSA-Nets, respectively.
(3)	By rotating and cyclic permutating the feature maps $F_{GE}$ and $F_{GSA}$, we obtained feature maps $F''_{GE}$ and $F''_{GSA}$ which are the ground truths of the feature maps of rotated image.
(4)	Finally, we got error maps $E_{GE} = F''_{GE}- F'_{GE}$, $E_{ GSA } = F''_{ GSA }- F'_{ GSA }$ as shown in Fig.~\ref{fig:errormap} which shows that our GE-ViT performs better.
The average errors of our GE-ViT are on the order of $10^{-5}$ while the average errors of GSA-Nets are on the order of $10^{-1}$.

\section{Discussion}\label{S7}

GE-ViT with a novel and effective positional encoding outperforms GSA-Nets and non-equivariant self-attention networks are competitive to G-CNNs. However, G-CNNS still performs better on most data sets, which may be due to the optimization problem of GE-ViT or the limits on computing resources. From the theoretical perspective, the group equivariant self-attention can be more expressive than G-CNNS, so the GE-ViT has a lot of potential for improvement in the aspect of initialization, optimization, generalization, etc.

\section*{Acknowledgements}
We gratefully thank the authors of GSA-Nets paper David W. Romero and Jean-Baptiste Cordonnier. They patiently answered and elaborated the experimental details of the paper GSA-Nets.

\balance
\bibliography{xu_298}
\balance
\appendix
\onecolumn

\section{Proof}\label{Appendix:mistake}

In this appendix, we prove that GE-ViT is group equivariant. For brevity, we also use the substitutions:
$$\bar{\gi} = x^{-1}(\bar{\gh}^{-1} (x(\gi) - y)) \Rightarrow \gi = x^{-1}(\bar{\gh}x(\bar{\gi}) + y)), \tilde{\gh}' = \bar{\gh}^{-1}\tilde{\gh},$$ 
and 
$$\bar{\gj} = x^{-1}(\bar{\gh}^{-1} (x(\gj) - y)) \Rightarrow \gj = x^{-1}(\bar{\gh}x(\bar{\gj}) + y)), \hat{\gh}' = \bar{\gh}^{-1}\hat{\gh}.$$

\subsection{Definitions and Notations.}

\subsubsection{Definition of Group Equivariant Self-Attention.} If the group self-attention formulation $\mathcal{m}^{r}_{\mathcal{G}}[f,\rho](i,\mathcal{h})$ is $\mathcal{G}$-equivariant, if and only if it satisfies: $$\mathcal{m}^{r}_\mathcal{G}[\mathcal{L}_{\mathcal{g}}[f],\rho](i,\mathcal{h})= \mathcal{L}_{\mathcal{g}}[\mathcal{m}^{r}_\mathcal{G}[f,\rho]](i,\mathcal{h}),\quad\mathcal{g}\in \mathcal{G}$$
\subsubsection{Input under $g$-Transformed}A $g$-transformed input can be expressed as:
\begin{align*}
    \mathcal{L}_{\mathcal{g}}[f](\mathcal{i}, \tilde{h}) = \mathcal{L}_{y}\mathcal{L}_{\bar{\mathcal{h}}}[f](\mathcal{i}, \tilde{\mathcal{h}}) = f(\rho^{-1}(\bar{\mathcal{h}}^{-1}(\rho(\mathcal{i}) - y)), \bar{\mathcal{h}}^{-1}\tilde{\mathcal{h}}),\\ 
    \mathcal{g} = (y, \bar{\mathcal{h}}),\; y \in \mathbb{R}^{d},\; \bar{\mathcal{h}} \in \mathcal{H}.
\end{align*}

\subsection{Proof of GE-ViT}

The complete proof process is as follows:
\begin{align}
    \mathcal{m}&^{r}_{\mathcal{G}}\big[\mathcal{L}_{y}\mathcal{L}_{\bar{\mathcal{h}}}[f], \rho\big](\mathcal{i}, \mathcal{h}) \\ \nonumber 
    &= \varphi_{\text{out}}\Big( \bigcup_{h \in [H]} \sum_{\tilde{\gh} \in \gH}\hspace{-1.05cm}  \sum_{\qquad \quad (\gj, \hat{\gh}) \in \gN(\gi, \tilde{\gh})} \hspace{-0.6cm}\hspace{-5mm}\sigma\hspace{-0.5mm}_{\gj, \hat{\gh}}\big(\langle \varphi_{\text{qry}}^{(h)}(\gL_{y}\gL_{\bar{\gh}}[f](\gi, \tilde{\gh})), \varphi_{\text{key}}^{(h)}(\gL_{y}\gL_{\bar{\gh}}[f](\gj, \hat{\gh})\\[-1\jot] \nonumber
    &\hspace{4.1cm} + \gL_{\gh}[\rho]((\gi, \tilde{\gh}), (\gj, \hat{\gh})) \rangle \big)
    \varphi_{\text{val}}^{(h)}(\gL_{y}\gL_{\bar{\gh}}[f](\gj, \hat{\gh})) \Big) \\
    &= \varphi_{\text{out}}\Big( \bigcup_{h \in [H]} \sum_{\tilde{\gh} \in \gH}\hspace{-1.05cm} \sum_{\qquad \quad (\gj, \hat{\gh}) \in \gN(\gi, \tilde{\gh})} \hspace{-0.6cm}\hspace{-5mm}\sigma\hspace{-0.5mm}_{\gj, \hat{\gh}}\big(\langle \varphi_{\text{qry}}^{(h)}(f(x^{-1}(\bar{\gh}^{-1}(x(\gi) - y)), \bar{\gh}^{-1}\tilde{\gh})),\\  \nonumber
    &\hspace{1cm}\varphi_{\text{key}}^{(h)}(f(x^{-1}(\bar{\gh}^{-1}(x(\gj) - y)), \bar{\gh}^{-1}\hat{\gh}) + \gL_{\gh}[\rho]((\gi, \tilde{\gh}), (\gj, \hat{\gh})) \rangle \big)\\ \nonumber
    &\hspace{2cm}\varphi_{\text{val}}^{(h)}(f(x^{-1}(\bar{\gh}^{-1}(x(\gj) - y)), \bar{\gh}^{-1}\hat{\gh})) \Big)\\ 
    &= \varphi_{\text{out}}\Big( \bigcup_{h \in [H]} \sum_{\bar{\gh}\tilde{\gh}' \in \gH} \hspace{-4.6cm}\sum_{\qquad\qquad\qquad\qquad\qquad\qquad \quad (x^{-1}(\bar{\gh}x(\bar{\gj}) + y), \bar{\gh}\hat{\gh}') \in \gN(x^{-1}(\bar{\gh}x(\bar{\gi}) + y), \bar{\gh}\tilde{\gh}')} \hspace{-4.4cm}\sigma\hspace{-0.5mm}_{x^{-1}(\bar{\gh}x(\bar{\gj}) + y), \bar{\gh}\hat{\gh}'}\big(\langle \varphi_{\text{qry}}^{(h)}(f(\bar{\gi}, \tilde{\gh}')),\varphi_{\text{key}}^{(h)}(f(\bar{\gj}, \hat{\gh}') \\
    \nonumber
    &\hspace{0.6cm} + \gL_{\textcolor{red}{\gh}}[\rho]((x^{-1}(\bar{\gh}x(\bar{\gi}) + y), \textcolor{red}{\bar{\gh}\tilde{\gh}'}),
    (x^{-1}(\bar{\gh}x(\bar{\gj}) + y), \textcolor{red}{\bar{\gh}\hat{\gh}'})) \rangle \big)\varphi_{\text{val}}^{(h)}(f(\bar{\gj}, \hat{\gh}')) \Big)
\end{align}
By using the definition:
$$
    \rho((i,\tilde{\mathcal{h}}), (j,\hat{\mathcal{h}})) = \rho^{P}(x(j)-x(i), \tilde{\mathcal{h}}\hat{\mathcal{h}}^{-1}\tilde{\mathcal{h}})
$$
and 
$$
    \mathcal{L}_{\mathcal{h}}[\rho]((i,\tilde{h}),(j,\hat{h})) = \rho^{P}(\mathcal{h}^{-1}(x(j)-x(i)),\mathcal{h}^{-1}(\tilde{\mathcal{h}}\hat{\mathcal{h}}^{-1}\tilde{\mathcal{h}})).
$$
The above formula can be further derived:
\begin{align}
& =\varphi_{\text{out}}\Big( \bigcup_{h \in [H]} \sum_{\bar{\gh}\tilde{\gh}' \in \gH} \hspace{-4.6cm}\sum_{\qquad\qquad\qquad\qquad\qquad\qquad \quad (x^{-1}(\bar{\gh}x(\bar{\gj}) + y), \bar{\gh}\hat{\gh}') \in \gN(x^{-1}(\bar{\gh}x(\bar{\gi}) + y), \bar{\gh}\tilde{\gh}')} \hspace{-4.4cm}\sigma\hspace{-0.5mm}_{x^{-1}(\bar{\gh}x(\bar{\gj}) + y), \bar{\gh}\hat{\gh}'}\big(\langle \varphi_{\text{qry}}^{(h)}(f(\bar{\gi}, \tilde{\gh}')), \varphi_{\text{key}}^{(h)}(f(\bar{\gj}, \hat{\gh}') \\[-0.5\jot]  \nonumber
&\hspace{0.35cm} + \rho^P( \gh^{-1}(\bar{\gh}x(\bar{\gj}) + y - (\bar{\gh}x(\bar{\gi}) + y)), \textcolor{red}{\gh^{-1}(\bar{\gh}\tilde{\gh}')(\bar{\gh}\hat{\gh}')^{-1}(\bar{\gh}\tilde{\gh}')} )) \rangle \big) \varphi_{\text{val}}^{(h)}(f(\bar{\gj}, \hat{\gh}')) \Big)\\ 
& =\varphi_{\text{out}}\Big( \bigcup_{h \in [H]} \sum_{\bar{\gh}\tilde{\gh}' \in \gH} \hspace{-4.6cm}\sum_{\qquad\qquad\qquad\qquad\qquad\qquad \quad (x^{-1}(\bar{\gh}x(\bar{\gj}) + y), \bar{\gh}\hat{\gh}') \in \gN(x^{-1}(\bar{\gh}x(\bar{\gi}) + y), \bar{\gh}\tilde{\gh}')} \hspace{-4.4cm}\sigma\hspace{-0.5mm}_{x^{-1}(\bar{\gh}x(\bar{\gj}) + y), \bar{\gh}\hat{\gh}'}\big(\langle \varphi_{\text{qry}}^{(h)}(f(\bar{\gi}, \tilde{\gh}')), \varphi_{\text{key}}^{(h)}(f(\bar{\gj}, \hat{\gh}') \\[-0.5\jot]  \nonumber
&\hspace{0.65cm} + \rho^P( \gh^{-1}(\bar{\gh}x(\bar{\gj}) + y - (\bar{\gh}x(\bar{\gi}) + y)), \textcolor{red}{\gh^{-1}\bar{\gh}\tilde{\gh}'{\hat{\gh}}^{'-1}\tilde{\gh}'} )) \rangle \big) \varphi_{\text{val}}^{(h)}(f(\bar{\gj}, \hat{\gh}')) \Big)\\
& = \varphi_{\text{out}}\Big( \bigcup_{h \in [H]} \sum_{\bar{\gh}\tilde{\gh}' \in \gH} \hspace{-4.6cm}\sum_{\qquad\qquad\qquad\qquad\qquad\qquad \quad (x^{-1}(\bar{\gh}x(\bar{\gj}) + y), \bar{\gh}\hat{\gh}') \in \gN(x^{-1}(\bar{\gh}x(\bar{\gi}) + y), \bar{\gh}\tilde{\gh}')} \hspace{-4.4cm}\sigma\hspace{-0.5mm}_{x^{-1}(\bar{\gh}x(\bar{\gj}) + y), \bar{\gh}\hat{\gh}'}\big(\langle \varphi_{\text{qry}}^{(h)}(f(\bar{\gi}, \tilde{\gh}')), \varphi_{\text{key}}^{(h)}(f(\bar{\gj}, \hat{\gh}')\\[-0.5\jot] \nonumber
&\hspace{3cm}  + \rho^P( \textcolor{red}{\gh^{-1}\bar{\gh}}(x(\bar{\gj})- x(\bar{\gi}), \textcolor{red}{\tilde{\gh}'{\hat{\gh}}^{'-1}\tilde{\gh}'}))) \rangle \big)  \varphi_{\text{val}}^{(h)}(f(\bar{\gj}, \hat{\gh}')) \Big) \\
& = \varphi_{\text{out}}\Big( \bigcup_{h \in [H]} \sum_{\bar{\gh}\tilde{\gh}' \in \gH} \hspace{-4.6cm}\sum_{\qquad\qquad\qquad\qquad\qquad\qquad \quad (x^{-1}(\bar{\gh}x(\bar{\gj}) + y), \bar{\gh}\hat{\gh}') \in \gN(x^{-1}(\bar{\gh}x(\bar{\gi}) + y), \bar{\gh}\tilde{\gh}')} \hspace{-4.4cm}\sigma\hspace{-0.5mm}_{x^{-1}(\bar{\gh}x(\bar{\gj}) + y), \bar{\gh}\hat{\gh}'}\big(\langle \varphi_{\text{qry}}^{(h)}(f(\bar{\gi}, \tilde{\gh}')), \varphi_{\text{key}}^{(h)}(f(\bar{\gj}, \hat{\gh}')  \\[-0.5\jot] \nonumber
&\hspace{3.65cm} + \gL_{\textcolor{red}{\bar{\gh}^{-1}\gh}}[\rho]((\bar{\gi}, \textcolor{red}{\tilde{\gh}'}),(\bar{\gj}, \textcolor{red}{\hat{\gh}'}))) \rangle \big)  \varphi_{\text{val}}^{(h)}(f(\bar{\gj}, \hat{\gh}')) \Big)
\end{align}

The subsequent proof is similar to the GSA-Nets~\citep{GSA-Nets}. For unimodular groups, the area of summation remains equal for any transformation $\gg \in \gG$, which means that:
\begin{align*} 
    \sum_{(x^{-1}(\bar{\gh}x(\bar{\gj}) + y), \bar{\gh}\hat{\gh}') \in \gN(x^{-1}(\bar{\gh}x(\bar{\gi}) + y), \bar{\gh}\tilde{\gh}')}  \hspace{-2cm} [\cdot] \hspace{1.5cm} \quad=& \hspace{0.5cm} \sum_{(x^{-1}(\bar{\gh}x(\bar{\gj})), \bar{\gh}\hat{\gh}') \in \gN(x^{-1}(\bar{\gh}x(\bar{\gi})), \bar{\gh}\tilde{\gh}')}  \hspace{-1.75cm} [\cdot] \\
    =& \hspace{0.8cm} \sum_{(x^{-1}(x(\bar{\gj})), \hat{\gh}') \in \gN(x^{-1}(x(\bar{\gi})), \tilde{\gh}')}  \hspace{-1.45cm} [\cdot] \hspace{1cm} \\ 
    =& \hspace{1.9cm}\sum_{(\bar{\gj}, \hat{\gh}') \in \gN(\bar{\gi}, \tilde{\gh}')} \hspace{-0.35cm} [\cdot].
\end{align*}
and because of the basic properties of groups, we can get $ \sum_{\bar{\gh}\tilde{\gh}' \in \gH}[\cdot] = \sum_{\tilde{\gh}' \in \gH}[\cdot]$. Consequently, the above formula can be further simplified as:

\begin{equation}
    \begin{aligned}
\gm^{r}_{\gG}\big[\gL_{\gy}\gL_{\bar{\gh}}[f], \rho\big](\gi, \gh) =& \varphi_{\text{out}}\Big( \bigcup_{h \in [H]}\sum_{\tilde{\gh}' \in \gH} \sum_{ (\bar{\gj}, \hat{\gh}') \in \gN(\bar{\gi}, \tilde{\gh}')} \hspace{-0.65cm}\sigma\hspace{-0.5mm}_{\bar{\gj}, \hat{\gh}'}\big(\langle 
\varphi_{\text{qry}}^{(h)}(f(\bar{\gi}, \tilde{\gh}')), 
 \\[-1\jot]
&\hspace{0.1cm}\varphi_{\text{key}}^{(h)}(f(\bar{\gj}, \hat{\gh}')  + 
\gL_{\bar{\gh}^{-1}\gh}[\rho]((\bar{\gi}, \tilde{\gh}'),(\bar{\gj}, \hat{\gh}'))) \rangle \big)  \varphi_{\text{val}}^{(h)}(f(\bar{\gj}, \hat{\gh}')) \Big)
\\[2\jot]
=&\gm^{r}_{\gG}[f, \rho](\bar{\gi}, \bar{h}^{-1}\gh) \\[2\jot]
=& \gm^{r}_{\gG}[f, \rho](x^{-1}(\bar{\gh}^{-1} (x(\gi) - y)), \bar{h}^{-1}\gh) 
\\[2\jot]
=& \gL_{y}\gL_{\bar{h}}\big[\gm^{r}_{\gG}[f, \rho]\big](\gi, \gh).
    \end{aligned}
\end{equation}
From the above formula, it can be seen that: 
$$
\gm^{r}_{\gG}[\gL_{y}\gL_{\bar{h}}[f], \rho](\gi, \gh) = \gL_{y}\gL_{\bar{h}}[\gm^{r}_{\gG}[f, \rho]](\gi, \gh),
$$
which is the same as:
$$
\mathcal{m}^{r}_\mathcal{G}[\mathcal{L}_{\mathcal{g}}[f],\rho](i,\mathcal{h})= \mathcal{L}_{\mathcal{g}}[\mathcal{m}^{r}_\mathcal{G}[f,\rho]](i,\mathcal{h}),\quad\mathcal{g}\in \mathcal{G}.
$$
Therefore, with the positional encoding we proposed:
$$
    \rho((i,\tilde{\mathcal{h}}), (j,\hat{\mathcal{h}})) = \rho^{P}(x(j)-x(i), \tilde{\mathcal{h}}\hat{\mathcal{h}}^{-1}\tilde{\mathcal{h}}),
$$
the group self-attention is group equivariant.

\end{document}